\documentclass{article}
\usepackage{spconf,amsmath,graphicx, multirow}

\title{Two-stream convolutional neural network for accurate RGB-D fingertip detection using depth and edge information}
\name{Hengkai Guo, Guijin Wang, Xinghao Chen}
\address{Department of Electronic Engineering\\
	Tsinghua University, Beijing}
\begin{document}
%
\maketitle
\begin{abstract}
Accurate detection of fingertips in depth image is critical for human-computer interaction. In this paper, we present a novel two-stream convolutional neural network (CNN) for RGB-D fingertip detection. Firstly edge image is extracted from raw depth image using random forest. Then the edge information is combined with depth information in our CNN structure. We study several fusion approaches and suggest a slow fusion strategy as a promising way of fingertip detection. As shown in our experiments, our real-time algorithm outperforms state-of-the-art fingertip detection methods on the public dataset HandNet with an average 3D error of 9.9mm, and shows comparable accuracy of fingertip estimation on NYU hand dataset.
\end{abstract}
\begin{keywords}
Convolutional Neural Network, Two-stream, Fingertip Detection, Edge detection, RGB-D imaging
\end{keywords}
\section{Introduction}
Hand pose estimation from depth image \cite{wang2013depth} \cite{yin2014efficient} is critical for human-computer interaction \cite{sharp2015accurate}, and has been studied extensively in recently years \cite{sun2015cascaded} \cite{li2015hand} \cite{tangopening} \cite{oberwegertraining}. Among all the joints of hand, fingertips play an important role in interaction, which are related to lots of gestures such as swipe and tap \cite{marin2014hand}. In the meanwhile, fingertip positions are often the most difficult to learn due to various hand poses, large self-occlusion and poor depth recovery near fingertips \cite{tompson2014real}. Most of existing approaches \cite{tang2014latent} \cite{sun2015cascaded} \cite{li2015hand} \cite{tangopening} \cite{oberwegertraining} rely on the topology structure of hand, leading to relative large error on fingertips (often larger than 1cm) due to error accumulation from palm to fingertip.

\begin{figure}[htb]
\centering
{\includegraphics[width=0.5\textwidth]{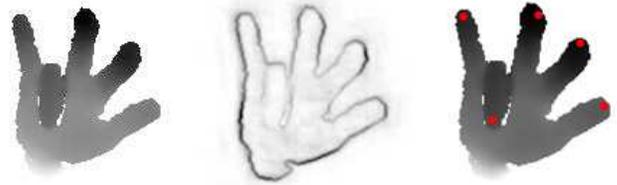}}
\caption{Depth image (left) and edge image (middle) of hand from HandNet\cite{wetzler2015rule} dataset. The red circles in the image (right) are the predicted fingertips with our algorithm.}
\label{fig_review}
\end{figure}

To solve these problems, we propose a new method for accurate detection of fingertip positions based on convolutional neural network (CNN). Different from earlier works that employ only depth image as input \cite{oberweger2015hands} \cite{tompson2014real} \cite{oberwegertraining}, we take advantage of both depth information and edge information from depth image (See Fig.\ref{fig_review}). We employ random forest (RF) to extract edges as in \cite{dollar2013structured}. Then we investigate a different architecture based on two separate streams with both depth and edge images, which are then combined by fusion strategy. After comparison with different CNN structures and fusion strategies, a deep structure with slow fusion is chosen for precise fingertip detection. We will show that such strategy is able to improve fingertip estimation performance. Evaluated on two public datasets \cite{wetzler2015rule}\cite{tompson2014real}, our method outperforms other state-of-the-art algorithms with a 3D error of 9.9mm on fingertips, and obtains comparable performance on the challenging NYU hand dataset.

\section{Related work}
We briefly review the previous works that are related to our work, including hand pose estimation, edge detection and CNN on hand pose and fingertip regression.

\noindent\textbf{Hand pose estimation.}\hspace{2mm} State-of-the-art algorithms can be roughly divided into two categories: generative methods and discriminative methods. Generative methods adopt particle swarm optimization (PSO) \cite{qian2014realtime} \cite{sharp2015accurate} or gradient descent \cite{sridhar2015fast} to optimize the energy function based on hand model. They all require initialization from either former frames \cite{sridhar2015fast} or coarse pose estimation \cite{qian2014realtime} \cite{sharp2015accurate}. Discriminative methods directly predict the positions of hand joints from images. Among them, RF is popular with hierarchical hand structure: from wrist to fingertip \cite{sun2015cascaded} \cite{tangopening}, or from whole to local \cite{tang2014latent} \cite{li2015hand}. While such methods are robust to complex articulations, they may suffer from error accumulation on fingertips. Our work focuses on more accurate fingertip detection for better human interaction with depth imaging \cite{shi2015high}.

\noindent\textbf{Edge information.}\hspace{2mm} Edges have been applied in high-level vision tasks such as object detection and object proposal generation \cite{qi2015making}, which can be easily extracted with forest \cite{dollar2013structured} or CNN \cite{xie2015holistically}. Edge information has also been implicitly used for extracting shape context feature in human pose estimation \cite{li2011local} \cite{he2015depth}. To the best of our knowledge, this is the first work to learn features from edges using CNN for joint regression.

\noindent\textbf{Convolutional neural network.}\hspace{2mm} Recently CNN structure has been employed for hand pose regression and fingertip detection. For hand pose estimation, Tompson et al. \cite{tompson2014real} use CNN to produce heat maps with the 2D joint positions, and then infer the 3D hand pose with inverse kinematics. Oberweger et al. \cite{oberweger2015hands} regress the hand pose with multi-scale and multi-stage CNN using pose prior. In \cite{oberwegertraining} three CNNs are used separately for initialization, generation of depth image and pose updating. For fingertip detection, Wetzler et al. \cite{wetzler2015rule} employ CNN for in-plane derotation of hand depth image and then use RF or CNN for fingertip estimation. In \cite{liu2015fingertip}, CNNs are designed to perform hand detection and index fingertip regression in egocentric RGB images. Different from previous CNN-based methods, our algorithm combines depth image with edge image using a two-stream CNN structure, which improves the accuracy of fingertip detection.

\section{Models}
\begin{figure}[htb]
\centering
{\includegraphics[width=0.5\textwidth]{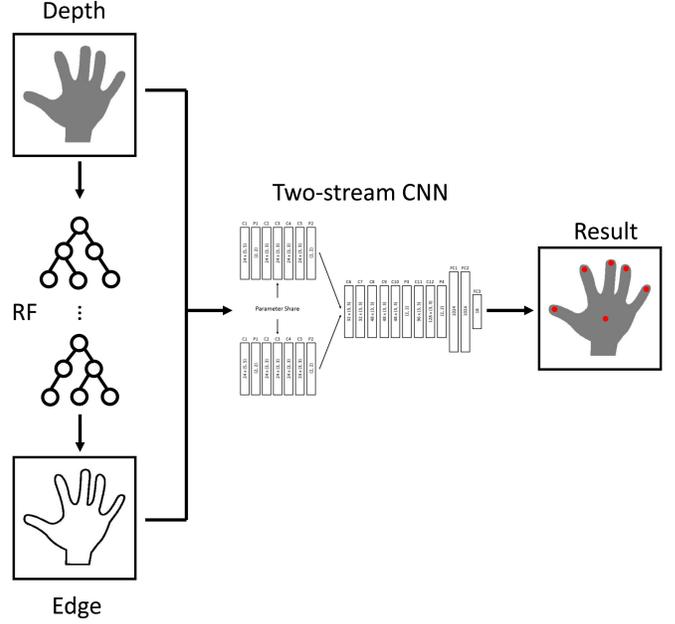}}
\caption{Overview of our algorithm. Edge image is detected from depth image using RF (left). Then both the depth and edge image are sent into two-stream CNN to obtain the palm and fingertip positions. }
\label{fig_algo}
\end{figure}

Fig. \ref{fig_algo} shows an overview of our algorithm. Given a depth image, hand region is first cropped and resized to $96\times96$ as in \cite{wetzler2015rule}. Then a pre-trained RF on NYU Depth dataset \cite{Silberman:ECCV12} is applied for edge detection from the cropped depth image (see \cite{dollar2013structured} for details of edge algorithm). Finally, a two-stream CNN regresses fingertip and palm positions using depth and edge information.

We next discuss the architecture of our two-stream CNN in details. This is followed by a description of algorithm implementation.

\subsection{Network Architecture}
To exploit the information of depth and edge, we investigate several strategies to fuse two-stream images, which are different in when to fuse. We first describe a baseline single-stream CNN and then discuss its extensions according to different types of fusion, including enhance fusion, early fusion, slow fusion, late fusion, and result fusion.

\begin{figure*}[htb]
\centering
\begin{minipage}[b]{0.48\textwidth}
  \centering
  \centerline{\includegraphics[width=0.8\textwidth]{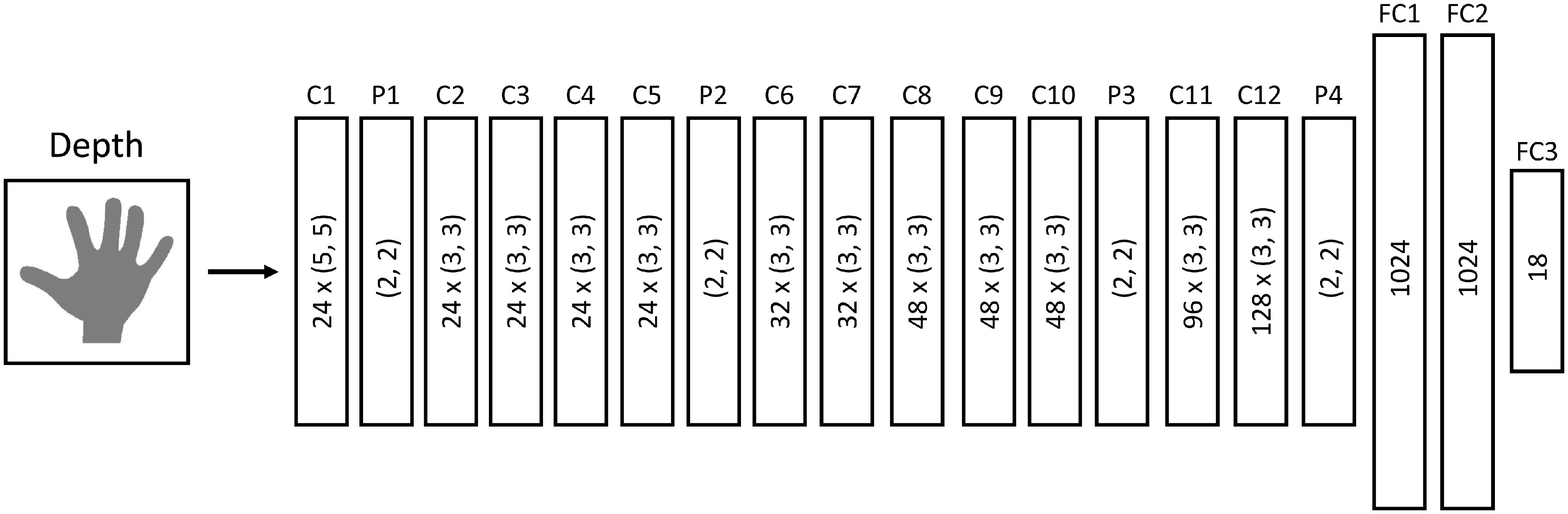}}
  \centerline{(a) Basic network}\medskip
\end{minipage}
\begin{minipage}[b]{0.48\textwidth}
  \centering
  \centerline{\includegraphics[width=0.8\textwidth]{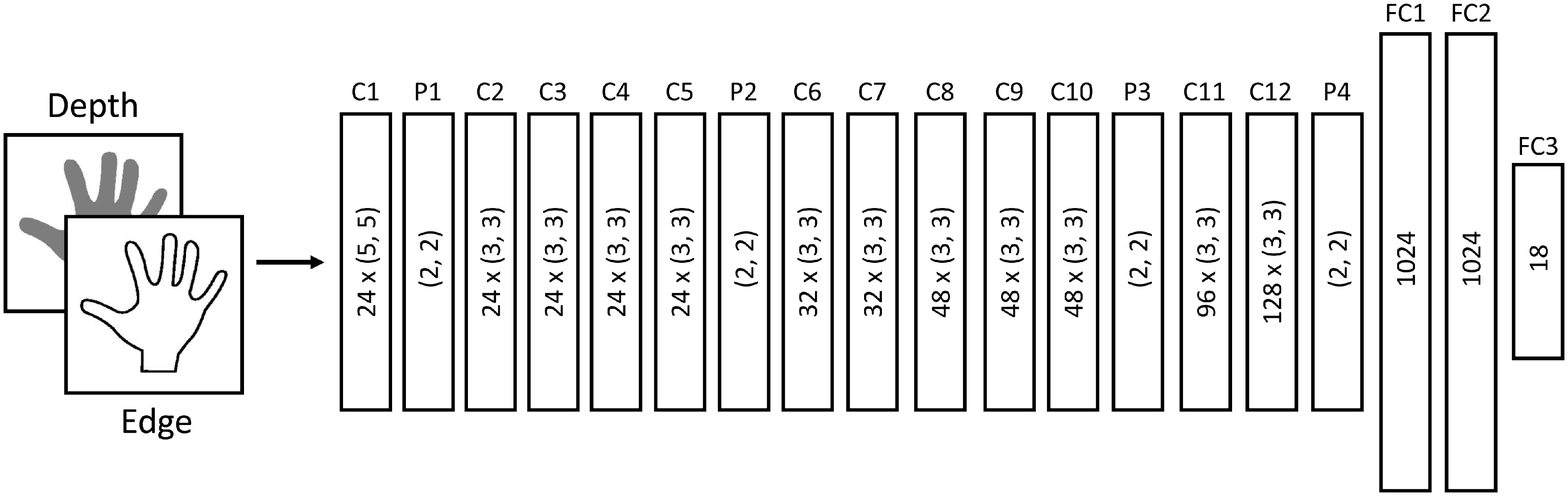}}
  \centerline{(b) Early fusion network}\medskip
\end{minipage}
\hfill
\begin{minipage}[b]{0.48\textwidth}
  \centering
  \centerline{\includegraphics[width=0.8\textwidth]{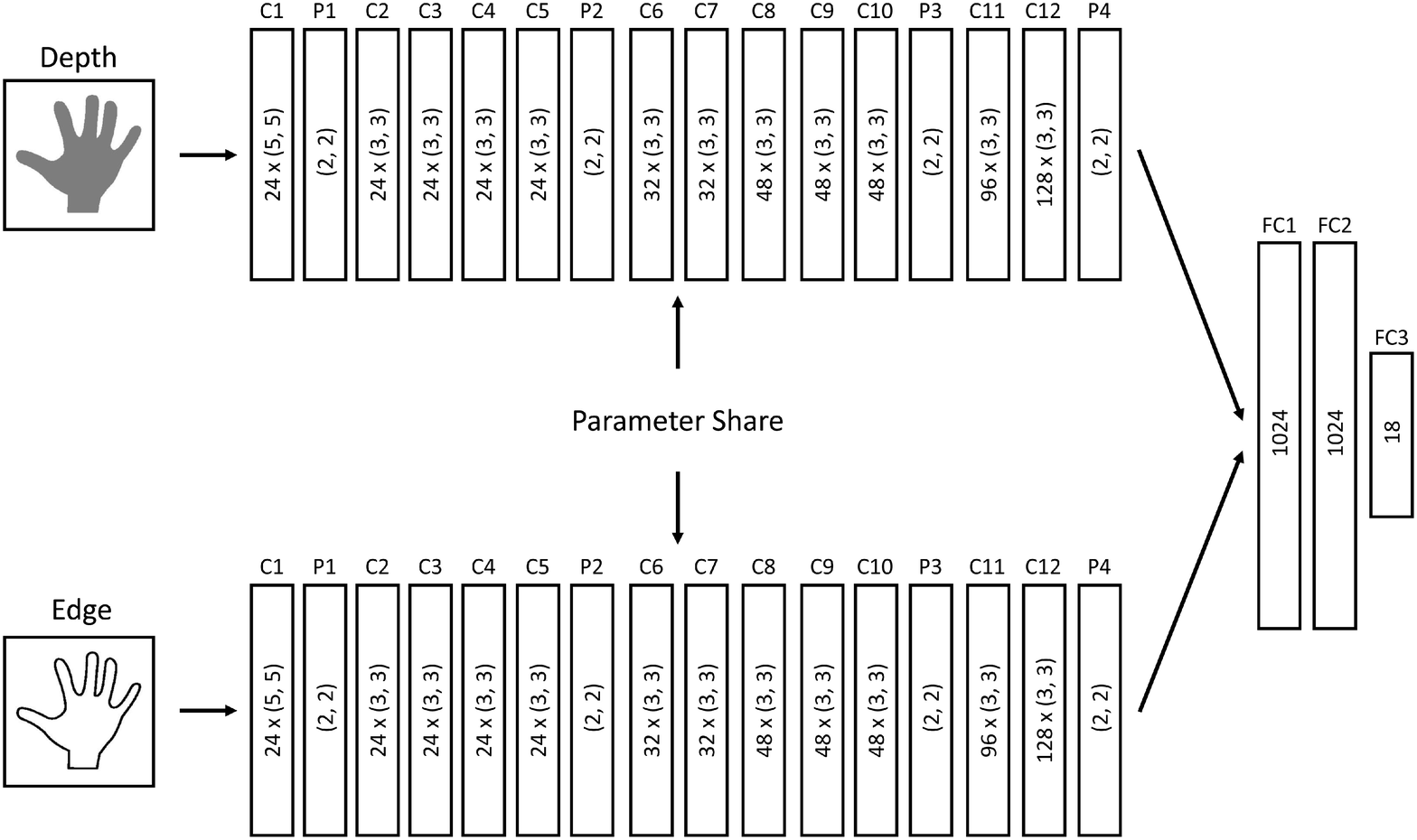}}
  \centerline{(c) Late fusion network}\medskip
\end{minipage}
\begin{minipage}[b]{0.48\textwidth}
  \centering
  \centerline{\includegraphics[width=0.8\textwidth]{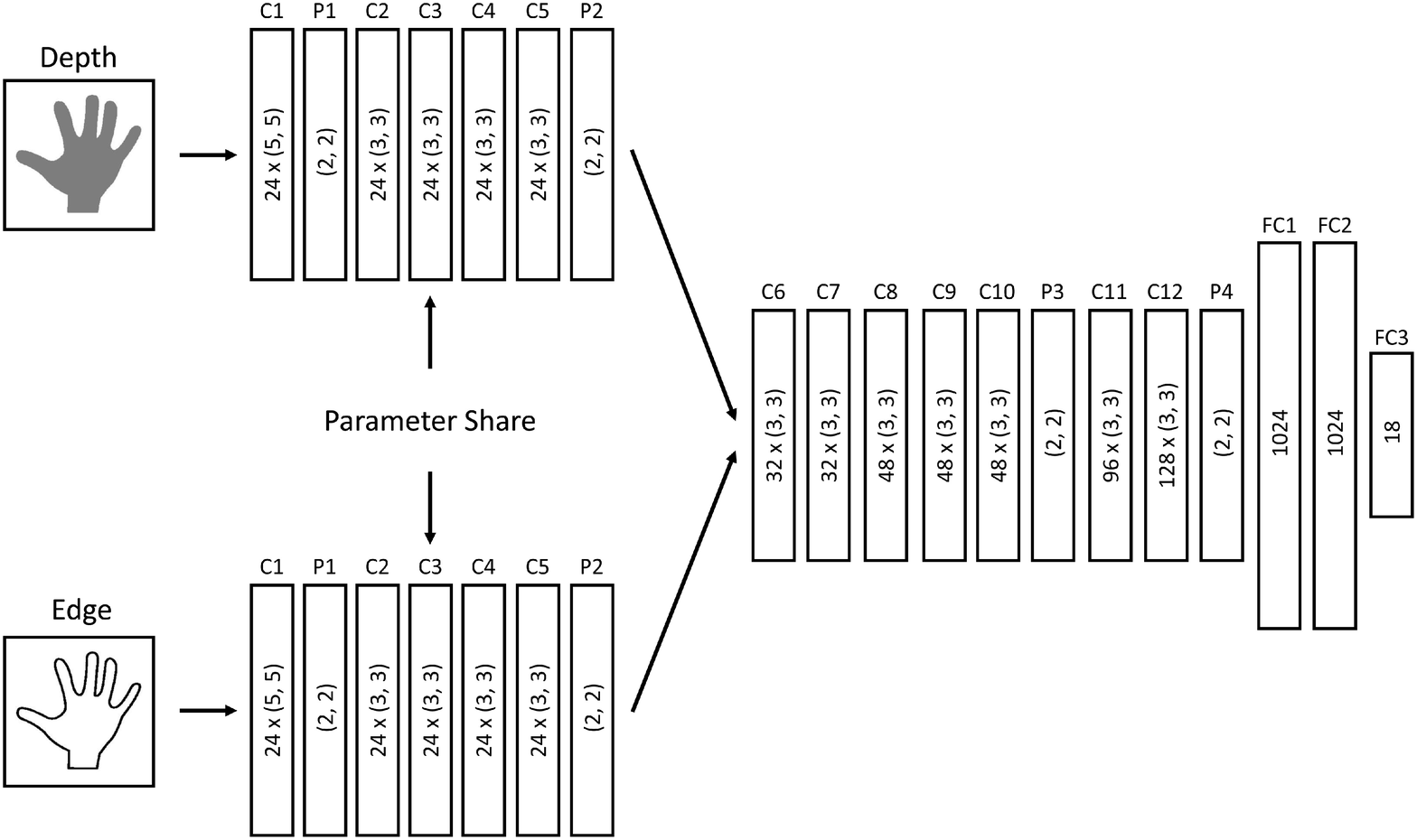}}
  \centerline{(d) Slow fusion network}\medskip
\end{minipage}
\caption{Different network structures for fingertip detection. $C$ denotes a convolutional layer with the number of filters and the filter size in the boxes, which is followed by a Rectified Linear Unit layer. $P$ denotes a max-pooling layer with the pooling size. $FC$ denotes a fully connected layer with the number of neurons.}
\label{fig_structure}
\end{figure*}

\noindent\textbf{Single-stream.}\hspace{2mm} We utilize a single-stream baseline architecture (Fig. \ref{fig_structure}(a)) to understand the contribution of each information. Using shorthand notation, the network is $C_1(24,5)-P_1-C_2(24,3)-C_3(24,3)-C_4(24,3)-C_5(24,3)-P_2-C_6(32,3)-C_7(32,3)-C_8(48,3)-C_9(48,3)-C_{10}(48,3)-P_3-C_{11}(96,3)-C_{12}(128,3) -P_4-FC_1(1024)-FC_2(1024)-FC_3(18)$, where $C(d,f)$ indicates a convolutional layer with $d$ filters of spatial size $f \times f$ when all the strides are 1. $FC(n)$ is a fully connected layer with $n$ nodes. All the pooling layers $P$ are non-overlapping max-pooling in $2 \times 2$ regions and each convolutional layer is followed by a Rectified Linear Unit (ReLU) layer. The output of $FC3$ is an 18-element vector with the 3D positions of fingertips and palm.

\noindent\textbf{Enhance fusion.}\hspace{2mm} The enhance fusion model shares the same structure with single-stream model, but with a different input formulated as:
\begin{equation}
I_{enhance} = 0.8I_{depth} + 0.2I_{edge}
\end{equation}
where $I_{depth}$ and $I_{edge}$ represent the depth image and edge image respectively. This can be seen as one kind of image sharpening, emphasizing the edges in depth image.

\noindent\textbf{Early fusion.}\hspace{2mm} The early fusion extension combines depth image and edge image immediately as a double-channel image, and also uses the same structure of single-stream model (Fig.\ref{fig_structure}(b)). The early concatenation allows the network to learn different low-level filters for each information and then fusion them for better estimation.

\noindent\textbf{Late fusion.}\hspace{2mm} The late fusion model employs two separate single-stream networks up to $C12$ layer with shared weights, and then merges them in the first fully connected layer (see Fig.\ref{fig_structure}(c)). Therefore, they share the same feature pattern to avoid overfitting for feature learning.

\noindent\textbf{Slow fusion.}\hspace{2mm} The slow fusion strategy is a balanced mixture between early fusion and late fusion. While sharing first five convolutional layers for low-level feature extraction, the two-stream networks are merged in $C6$ layer for high-level fusion as in Fig.\ref{fig_structure}(d).

\noindent\textbf{Result fusion.}\hspace{2mm} The result fusion model simply applies two separate single-stream networks on depth image and edge image without shared parameters. The predicted positions are averaged for final results.

\subsection{Implementation details}
For edge detection, we adopt the public Matlab implementation in \cite{dollar2013structured} with their pre-trained depth model. For fingertip detection, we use Caffe \cite{jia2014caffe} to train the CNNs on a Nvidia Titan X GPU. We train the different architectures by minimizing the Euclidean loss between the prediction and ground truth using a batch size of 196 and learning rate of 0.01. All the networks converge within 400000 iterations. Similar to \cite{wetzler2015rule}, we do not perform any data augmentation during training.

\section{Results}
Experiments in this work are conducted on the HandNet fingertip dataset \cite{wetzler2015rule} and NYU Hand dataset \cite{tompson2014real}, following the settings in \cite{wetzler2015rule}. Mean precision (mP) as defined in \cite{wetzler2015rule} and mean 3D error of fingertips ($err_{f}$) is evaluated for each algorithm. Because no confidence score is predicted from our method, we do not calculate mAP as in \cite{wetzler2015rule}. Running time is also computed in the same environment with an Intel i7-4790 CPU, which has considered the time for edge detection. We first investigate different designs of network architecture on the HandNet dataset. Then we compare with the state-of-the-art methods on both datasets to show the efficiency of our algorithms.

\begin{table}[htb]
\label{table_result}
\caption{Result of different methods on the HandNet\cite{wetzler2015rule} test set. mP values indicate the mean precision defined in \cite{wetzler2015rule}. $err_{f}$ means the average 3D error of fingertips. Time indicates the running time for single image. }
\centering
\begin{tabular}{|c|c|c|c|}
\hline
Method & mP & $err_{f}$/mm & time/ms \\
\hline
CNN-depth-shallow & 0.744 & 11.9 & \textbf{4.98} \\
CNN-depth-median & 0.775 & 11.1 & 6.54 \\
CNN-depth-deep & 0.816 & 10.0 & 8.02 \\
CNN-edge-deep & 0.562 & 16.6 & 31.25 \\
CNN-depth-deep-finger & 0.803 & 10.4 & 8.05 \\
\hline
CNN-fusion-enhance & 0.786 & 10.8 & 32.34 \\
CNN-fusion-early & 0.772 & 11.5 & 32.26 \\
CNN-fusion-late & 0.801 & 10.6 & 50.21 \\
CNN-fusion-result & 0.776 & 11.1 & 33.32 \\
\textbf{CNN-fusion-slow} & \textbf{0.820} & \textbf{9.9} & 37.04 \\
\hline
\end{tabular}
\end{table}

\noindent\textbf{Network depth.}\hspace{2mm} To study the effect of network depth, two single-stream networks with shallow and median depth (CNN-depth-shallow and CNN-depth-median) are involved with the similar architecture of single-stream baseline CNN (CNN-depth-deep). The median CNN removes three layers C5, C10 and C12 comparing with the deep one. And the shallow CNN further takes off two layers C11 and P4 when changing the stride of C8 to 2 with to the median one. The results (the first three rows in the second group in Table. \ref{table_result}) show that the performance can be significantly improved by including more layers (1.9mm error dropped from shallow CNN to deep CNN), because deeper structures help to form high-level features and obtain a larger receptive fields on original image.

\begin{figure}[htb]
\centering
{\includegraphics[width=0.5\textwidth]{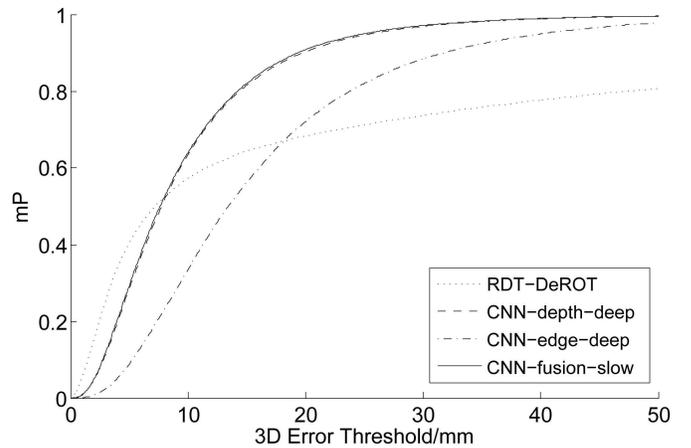}}
\caption{Mean precision under different thresholds for proposed method and other algorithms.}
\label{fig_res}
\end{figure}

\noindent\textbf{Contributions of edges and palm position.}\hspace{2mm} Single-stream depth-based CNN (CNN-depth-deep) and edge-based CNN (CNN-edge-deep) are implemented to investigate the importance of different information sources. A CNN with the same architecture but regresses only fingertip positions is also included to study the influence of palm position prediction. From Table. 1 and Fig. \ref{fig_res}, our two-stream slow-fusion CNN (CNN-fusion-slow) outperforms all the single-stream architectures on mP and mean error of fingertips. And there is a wide gap between depth-based and edge-based single-stream CNNs, because of the information loss from raw images to edges. Surprisingly, we observe an error drop (0.4mm) when training to predict only fingertip positions without palm. While detecting the palm may add slight burden for the network, the palm position itself is a strong prior to the fingertips, which is learned by CNN for better regression of fingertips.

\noindent\textbf{Fusion strategies.}\hspace{2mm} Different fusion approaches are also compared in our experiments. The slow-fusion structure achieves the highest mP and the lowest mean error among all the methods as shown in the last group in Table. \ref{table_result}, which indicates that the edge image and depth image may share similar low-level features and have diverse high-level features. So too early or late for feature fusion may deteriorate the performance of networks.

\noindent\textbf{Comparison with state-of-the-arts.}\hspace{2mm} Table.2 shows the results. For HandNet dataset, we compare our methods with random decision tree (RDT) and CNN algorithms in \cite{wetzler2015rule} on their provided results. Our methods outperform all the state-of-the-art approaches with a large margin in mP from 0.63 (RDT) or 0.61 (CNN) to 0.82 when no data augmentation is applied as in \cite{wetzler2015rule}. Moreover, the slow-fusion structure obtains an average 3D error of 9.9mm on fingertip prediction, which is nearly half of that in \cite{wetzler2015rule} and is smaller than the distance between adjacent fingertips \cite{wetzler2015rule}. For NYU dataset, our two-stream method is better than all the other CNN-based algorithms with 19.3mm errors on fingertips. However, RDT seems to handle such dataset better than CNNs. This may because the dataset enrolls single subject for training and two another subjects for testing, which makes CNN easier to overfit on training data. 

\begin{table}[htb]
\label{table_result2}
\caption{Result of different methods on the HandNet\cite{wetzler2015rule} and NYU \cite{tompson2014real} test set. mP values indicate the mean precision defined in \cite{wetzler2015rule}. $err_{f}$ means the average 3D error of fingertip (for methods in \cite{wetzler2015rule}. we discard all the undefined results whose errors are larger than 30cm).}
\centering
\begin{tabular}{|c|c|c|c|}
\hline
Dataset & Method & mP & $err_{f}$/mm \\
\hline
\multirow{3}{*}{HandNet} & RDT-DeROT\cite{wetzler2015rule} & 0.63 & 18.6 \\
& CNN-DeROT\cite{wetzler2015rule} & 0.61 & 23.4 \\
& CNN-fusion-slow & \textbf{0.82} & \textbf{9.9} \\
\hline
\multirow{5}{*}{NYU} & RDT-DeROT\cite{wetzler2015rule} & \textbf{0.63} & - \\
& CNN-DeROT\cite{wetzler2015rule} & 0.49 & - \\
& CNN-DeepPrior\cite{oberweger2015hands} & 0.43 & 26.4 \\
& CNN-Feedback\cite{oberwegertraining} & 0.38 & 23.2 \\
& CNN-fusion-slow & 0.50 & \textbf{19.3} \\
\hline
\end{tabular}
\end{table}

\noindent\textbf{Timing.} In Table. 1, all our methods provide acceptable speed on CPU for real-time fingertip detection. While the single-stream depth-based CNN performs up to over 125 fps, the two-stream slow-fusion architecture reaches 25 fps, which is enough for real-time application.

\section{Conclusion}
We propose an effective two-stream convolutional network for RGB-D fingertip detection. Edge information is slowly fused with depth information to achieve better accuracy for regression. By evaluation on HandNet dataset, we found that deep structure, two-stream information, slow-fusion strategy and palm constraint are the key factors for precise fingertip estimation. Our method significantly improves the prediction performance of state-of-the-art methods. Next we will apply the same idea on full hand pose estimation \cite{oberwegertraining} or human pose estimation \cite{he2015depth}. Learning when to fuse instead of manual design is also a promising direction for our work.

\noindent\textbf{Acknowledgments.}\hspace{2mm} This work is supported by The National Science Foundation of China (No. 61271390), and State High-Tech Development Plan (No. 2015AA016304).


\bibliographystyle{IEEEbib}
\bibliography{refs}

\begin{thebibliography}{10}

\bibitem{wang2013depth}
Guijin Wang, Xuanwu Yin, Xiaokang Pei, and Chenbo Shi,
\newblock ``Depth estimation for speckle projection system using progressive
  reliable points growing matching,''
\newblock {\em Applied optics}, vol. 52, no. 3, pp. 516--524, 2013.

\bibitem{yin2014efficient}
Xuanwu Yin, Guijin Wang, Chenbo Shi, and Qingmin Liao,
\newblock ``Efficient active depth sensing by laser speckle projection
  system,''
\newblock {\em Optical Engineering}, vol. 53, no. 1, pp. 013105--013105, 2014.

\bibitem{sharp2015accurate}
Toby Sharp, Cem Keskin, Duncan Robertson, Jonathan Taylor, Jamie Shotton, David
  Kim Christoph Rhemann~Ido Leichter, Alon Vinnikov~Yichen Wei, Daniel Freedman
  Pushmeet Kohli~Eyal Krupka, Andrew Fitzgibbon, and Shahram Izadi,
\newblock ``Accurate, robust, and flexible real-time hand tracking,''
\newblock in {\em Proc. CHI}, 2015, vol.~8.

\bibitem{sun2015cascaded}
Xiao Sun, Yichen Wei, Shuang Liang, Xiaoou Tang, and Jian Sun,
\newblock ``Cascaded hand pose regression,''
\newblock in {\em Proceedings of the IEEE Conference on Computer Vision and
  Pattern Recognition}, 2015, pp. 824--832.

\bibitem{li2015hand}
Peiyi Li, Haibin Ling, Xi~Li, and Chunyuan Liao,
\newblock ``3d hand pose estimation using randomized decision forest with
  segmentation index points,''
\newblock in {\em European Conference on Computer Vision}, 2015.

\bibitem{tangopening}
Danhang Tang, Jonathan Taylor, Pushmeet Kohli, Cem Keskin, Tae-Kyun Kim, and
  Jamie Shotton,
\newblock ``Opening the black box: Hierarchical sampling optimization for
  estimating human hand pose,''
\newblock in {\em European Conference on Computer Vision}, 2015.

\bibitem{oberwegertraining}
Markus Oberweger, Paul Wohlhart, and Vincent Lepetit,
\newblock ``Training a feedback loop for hand pose estimation,''
\newblock in {\em European Conference on Computer Vision}, 2015.

\bibitem{marin2014hand}
Giulio Marin, Fabio Dominio, and Pietro Zanuttigh,
\newblock ``Hand gesture recognition with leap motion and kinect devices,''
\newblock in {\em Image Processing (ICIP), 2014 IEEE International Conference
  on}. IEEE, 2014, pp. 1565--1569.

\bibitem{tompson2014real}
Jonathan Tompson, Murphy Stein, Yann Lecun, and Ken Perlin,
\newblock ``Real-time continuous pose recovery of human hands using
  convolutional networks,''
\newblock {\em ACM Transactions on Graphics (TOG)}, vol. 33, no. 5, pp. 169,
  2014.

\bibitem{tang2014latent}
Danhang Tang, Hyung~Jin Chang, Alykhan Tejani, and Tae-Kyun Kim,
\newblock ``Latent regression forest: Structured estimation of 3d articulated
  hand posture,''
\newblock in {\em Computer Vision and Pattern Recognition (CVPR), 2014 IEEE
  Conference on}. IEEE, 2014, pp. 3786--3793.

\bibitem{oberweger2015hands}
Markus Oberweger, Paul Wohlhart, and Vincent Lepetit,
\newblock ``Hands deep in deep learning for hand pose estimation,''
\newblock {\em Computer Vision Winter Workshop}, 2015.

\bibitem{dollar2013structured}
Piotr Doll{\'a}r and C~Lawrence Zitnick,
\newblock ``Structured forests for fast edge detection,''
\newblock in {\em Computer Vision (ICCV), 2013 IEEE International Conference
  on}. IEEE, 2013, pp. 1841--1848.

\bibitem{wetzler2015rule}
Aaron Wetzler, Ron Slossberg, and Ron Kimmel,
\newblock ``Rule of thumb: Deep derotation for improved fingertip detection,''
\newblock in {\em British Machine Vision Conference}, 2015.

\bibitem{qian2014realtime}
Chen Qian, Xiao Sun, Yichen Wei, Xiaoou Tang, and Jian Sun,
\newblock ``Realtime and robust hand tracking from depth,''
\newblock in {\em Computer Vision and Pattern Recognition (CVPR), 2014 IEEE
  Conference on}. IEEE, 2014, pp. 1106--1113.

\bibitem{sridhar2015fast}
Srinath Sridhar, Franziska Mueller, Antti Oulasvirta, and Christian Theobalt,
\newblock ``Fast and robust hand tracking using detection-guided
  optimization,''
\newblock in {\em Proceedings of the IEEE Conference on Computer Vision and
  Pattern Recognition}, 2015, pp. 3213--3221.

\bibitem{shi2015high}
Chenbo Shi, Guijin Wang, Xuanwu Yin, Xiaokang Pei, Bei He, and Xinggang Lin,
\newblock ``High-accuracy stereo matching based on adaptive ground control
  points,''
\newblock {\em Image Processing, IEEE Transactions on}, vol. 24, no. 4, pp.
  1412--1423, 2015.

\bibitem{qi2015making}
Yonggang Qi, Yi-Zhe Song, Tao Xiang, Honggang Zhang, Timothy Hospedales, Yi~Li,
  and Jun Guo,
\newblock ``Making better use of edges via perceptual grouping,''
\newblock in {\em Proceedings of the IEEE Conference on Computer Vision and
  Pattern Recognition}, 2015, pp. 1856--1865.

\bibitem{xie2015holistically}
Saining Xie and Zhuowen Tu,
\newblock ``Holistically-nested edge detection,''
\newblock {\em arXiv preprint arXiv:1504.06375}, 2015.

\bibitem{li2011local}
Zhenning Li and Dana Kuli{\'c},
\newblock ``Local shape context based real-time endpoint body part detection
  and identification from depth images,''
\newblock in {\em Computer and Robot Vision (CRV), 2011 Canadian Conference
  on}. IEEE, 2011, pp. 219--226.

\bibitem{he2015depth}
Li~He, Guijin Wang, Qingmin Liao, and Jing-Hao Xue,
\newblock ``Depth-images-based pose estimation using regression forests and
  graphical models,''
\newblock {\em Neurocomputing}, vol. 164, pp. 210--219, 2015.

\bibitem{liu2015fingertip}
Xiaorui Liu, Yichao Huang, Xin Zhang, and Lianwen Jin,
\newblock ``Fingertip in the eye: A cascaded cnn pipeline for the real-time
  fingertip detection in egocentric videos,''
\newblock {\em arXiv preprint arXiv:1511.02282}, 2015.

\bibitem{Silberman:ECCV12}
Pushmeet~Kohli Nathan~Silberman, Derek~Hoiem and Rob Fergus,
\newblock ``Indoor segmentation and support inference from rgbd images,''
\newblock in {\em ECCV}, 2012.

\bibitem{jia2014caffe}
Yangqing Jia, Evan Shelhamer, Jeff Donahue, Sergey Karayev, Jonathan Long, Ross
  Girshick, Sergio Guadarrama, and Trevor Darrell,
\newblock ``Caffe: Convolutional architecture for fast feature embedding,''
\newblock {\em arXiv preprint arXiv:1408.5093}, 2014.

\end{thebibliography}

\end{document}